\pdfoutput=1
\documentclass[11pt]{article}

\usepackage{ACL2023}

\usepackage{times}
\usepackage{latexsym}

\usepackage[T1]{fontenc}
\usepackage[utf8]{inputenc}

\usepackage{microtype}

\usepackage{inconsolata}

\usepackage{url}
\usepackage{booktabs}
\usepackage{graphicx}
\usepackage{xspace}

\newcommand{\eg}{\textit{e.g.,}\xspace}
\newcommand{\fen}{F\textsubscript{1}\xspace}
\title{UIO at SemEval-2023 Task 12:  Multilingual fine-tuning for sentiment classification in low-resource languages}

\author{Egil Rønningstad \\
  University of Oslo \\
  Department of Informatics \\
  \texttt{egilron@ifi.uio.no} 
}

\begin{document}
\maketitle
\begin{abstract}
Our contribution to the 2023 AfriSenti-SemEval shared task 12:  Sentiment Analysis for African Languages, provides insight into how a  multilingual large language model can be a resource for sentiment analysis in languages not seen during pretraining. The shared task provides datasets of a variety of African languages from different language families. The languages are to various degrees related to languages used during pretraining, and the language data contain various degrees of code-switching.  We experiment with both monolingual and multilingual datasets for the final fine-tuning, and find that with the provided datasets that contain samples in the thousands, monolingual fine-tuning yields the best results.
\end{abstract}

\section{Introduction}
The 2023 AfriSenti-SemEval Shared Task 12 is the first SemEval shared task for sentiment analysis, targeting African low-resource languages \citep{muhammadSemEval2023}. It aims to raise awareness for the need of annotated data in languages that receive little attention when it comes to building AI tools for the digital world.  

The task is, for each tweet in the dataset, to classify them correctly as conveying a negative, neutral or positive sentiment. This task of classifying sentiment category for microblog statements or individual sentences is a useful component in various Natural Language Processing (NLP) tasks. The problem is well researched for English, where similar tasks are modelled with more than 97\% accuracy.\footnote{\url{https://paperswithcode.com/task/sentiment-analysis}}

The shared task at hand is split in subtask A, B and C, where subtask A provides training data for 12 African languages, and subtask B provides a joint, multilingual train set for the same languages. We did not participate in subtask C which provided test data in two languages for Zero-shot inference.

Our work shares insight to the effect of fine-tuning a multilingual large language model (llm), for  languages not seen during pretraining. We compare the performance of the resulting model with its exposure to similar languages during the various steps of training and fine-tuning. We find that models fine-tuned on larger training sets, and models fine-tuned on languages close to those seen during pretraining and initial fine-tuning, perform the best. We found the "XLM-Twitter-sentiment" model presented in Section \ref{sec:plm} to be the best starting point according to our constraints. This model is both adapted to multilingual Twitter data in pretraining, and is initially fine-tuned to sentiment classification on a multilingual Twitter sentiment classification dataset. 

We fine-tuned this model on the provided monolingual data, and compared this with alternative models fine-tuned on the the multilingual dataset containing all 12 languages, and also models trained on a concatenation of the training data for the languages in the same language family. This work is presented in Section \ref{seq:ourwork}. We found that for the given resources in this task, monolingual fine-tuning yielded overall best results.

\section{Background}
Sentiment analysis provides insight into opinions and moods held in the population that the authors of the texts represent \citep{agarwal-etal-2011-sentiment, liu_2017}. It may also be an embedded part of a Natural Language Processing (NLP) pipeline, where the end result may be, \eg a dialogue system or an analysis of customer satisfaction. Sentiment analysis can be performed on various levels, and classifying texts into the categories of "positive", "neutral" or "negative" is not particularly fine-grained. However, this granularity can be modelled with high accuracy, in particular for well-resourced languages. Short texts like Twitter-messages are relatively easy to classify as they are often times opinionated, and may often express only one sentiment.

% an important sentiment analysis task. There are unique challenges involved in classifying the sentiment of longer texts, where multiple opinions may be expressed, and there may be long-range dependencies within the text that are difficult for neural models to capture. 

% SLutter for brått.

\subsection{Previous multilingual sentiment analysis tasks}

There has been a number of shared sentiment analysis tasks at SemEval earlier. The data have mainly been on the major languages of the world, and on various European languages. Three recent examples are:
\begin{itemize}
    \item SemEval 2022 Task 10: Structures sentiment analysis, utilizing  Norwegian, Basque, Catalan, Spanish and English data
    \item SemEval 2020 Task9: SentiMix, English-Hindi and English-Spanish code-mixed 
    data
    \item SemEval 2017 Task 4: Sentiment Analysis in Twitter, Arabic and English
\end{itemize}

\section{Pretrained language models}\label{sec:plm}
Fine-tuning an already pretrained llm can be seen as the de-facto standard approach to NLP tasks of sentiment analysis. We decided to search for one multilingual llm that could provide good results for all languages in the competition. We have experienced models based on xlm-Roberta (XLM-R) by \citet{conneau-etal-2020-unsupervised} to be a good starting point for multilingual sentiment analysis. For the low-resource languages in the shared task, we are not aware of any single model that is pretrained on all the languages in the competition, but the  \textbf{AfroXLMR} \citep{alabi-etal-2022-adapting} is pretrained on a majority of the included languages. As far as we understand, Hausa, Amharic, Arabic, Swahili and Portugese were present in the training data for both XLM-R and AfroXLMR. Yoruba, Igbo and   Kinyarwanda were present in the pretraining of AfroXLMR, but not in XLM-R. 

As llms may suffer not only from language barriers, but also from domain barriers  \citep{aue2005customizing}, we found a recent version of XLM-R to be of particular interest; the \textbf{XLM-Twitter}\xspace (XLM-T) model by \citet{barbieri-etal-2022-xlm}. The model is a result of further pretraining of XLM-R models on twitter data (198M tweets, 12G of uncompressed text). The twitter data were not filtered according to language. English, Portugese and Arabic are all among the top four best represented languages. Amharic is within the top 30 best represented languages in their additional pretraining on Twitter data. Further details on this model are presented in Section \ref{sec:chosen_model}, where we also present the \textbf{XLM-Twitter-sentiment} (XLMT-sentiment)\footnote{\url{cardiffnlp/twitter-xlm-roberta-base-sentiment}} model which comes already fine-tuned on a multilingual Twitter sentiment dataset.

We included an \textbf{mpnet-model} \citep{reimers-gurevych-2019-sentence} for comparison, since we consider the concept of sentence-transformers to be relevant to this task. Its performance was on par with the competition for some languages, and is an interesting approach worthy of further studies. The model was the best for Nigerian Pidgin, but had not strong enough overall performance across the languages. 

The above mentioned models were fine-tuned and evaluated on the shared task data for each language in subtask A. The results are reported in Table \ref{tab:init_models}, and we decided to use the XLMT-sentiment model as the pretrained llm for all our further experiments.

\begin{table*}
\begin{tabular}{lrrrr}
\toprule

\textbf{Model} & \textbf{afro-xlmr-} & \textbf{mpnet-} & \textbf{XLM-Twitter-} & \textbf{XLMT-sentiment-} \\
\textbf{Language} & \textbf{mini} & \textbf{base-v2} & \textbf{base}& \textbf{base}\\
\midrule
Amharic & 58.5\% & 45.0\% & 58.5\% & 63.5\% \\
Algerian Arabic & 64.0\% & 57.5\% & 66.5\% & 68.0\% \\
Hausa  & 74.5\% & 71.5\% & 75.0\% & 71.5\% \\
Igbo & 74.0\% & 72.5\% & 74.5\% & 75.0\% \\
Kinyarwanda & 60.0\% & 59.0\% & 63.5\% & 63.0\% \\
Moroccan Arabic(Darija), & 75.5\% & 70.0\% & 81.5\% & 78.0\% \\
Nigerian Pidgin & 72.0\% & 78.0\% & 77.0\% & 74.0\% \\
Mozambican Portuguese & 62.0\% & 59.5\% & 72.0\% & 70.0\% \\
Swahili & 57.0\% & 57.5\% & 58.5\% & 58.5\% \\
Xitsonga & 49.5\% & 47.5\% & 55.0\% & 58.5\% \\
Twi & 62.0\% & 65.0\% & 65.5\% & 68.0\% \\
Yoruba & 73.5\% & 74.0\% & 79.0\% & 75.5\% \\
\midrule % \\wsl$\Ubuntu-20.04\home\egil\gits_wsl\afrisent\results\pred_200\model_evals.xlsx
Mean & 65.2\% & 63.1\% & 68.9\% & 68.6\% \\
Lowest & 49.5\% & 45.0\% & 55.0\% & 58.5\% \\
\bottomrule
\end{tabular}
\caption{Initial results (\fen) from fine-tuning four pretrained llms on each language individually, and testing on a dev split created from the initial training data. Although XLM-Twitter had the highest average scores, we chose XLMT-sentiment for our contribution, since it was noticeably better on the weakest language.  }
\label{tab:init_models}
\end{table*}

\section{Dataset}   
We trained our model on only the data provided by the shared task. The twelve languages in the training dataset are represented with annotated tweets counting from 804 to 14172 in the training split, as can be seen in Table \ref{tab:languages}. The dataset  by \citet{muhammad2023afrisenti} builds on the work of \citet{muhammad-etal-2022-naijasenti} and \citet{yimam-etal-2020-exploring}. The dataset includes two Creole languages,  Nigerian Pidgin and Mozambican Portuguese, and two arabic languages, Algerian Arabic and Moroccan Arabic / Darija. In addition there is an amount of code-switching in the data \citep{muhammad-etal-2022-naijasenti}. The languages have therefore various levels of similarity, shared vocabulary or closeness to larger languages that our llm was pretrained on.

In addition to the training data for each language, the task includes a pre-shuffled dataset containing data from all the individual language datasets, for the multilingual Task B.

\begin{table*} % report_results.ipynb
\begin{tabular}{lllr}
\toprule
\textbf{Symbol} &                \textbf{Language} &                                     \textbf{Family} &  \textbf{Train} \\

\midrule
am       &                 Amharic &  Afro-Asiatic, Semitic, South, Ethiopian  &   5984 \\
dz       &          Algerian Arabic &      Afro-Asiatic, Semitic,  Arabic &   1651 \\
ha       &                  Hausa  &                 Afro-Asiatic, Chadic, West  &  14172 \\
ig       &                    Igbo &  Niger-Congo, Atlantic-Congo, Volta-Congo  &  10192 \\
kr       &             Kinyarwanda &  Niger-Congo, Atlantic-Congo, Volta-Congo  &   3302 \\
ma       &  Moroccan Arabic / Darija, &      Afro-Asiatic, Semitic, Arabic &   5583 \\
pcm      &         Nigerian Pidgin &                              Creole, English based  &   5121 \\
pt       &    Mozambican Portuguese &                           Creole, Portuguese based &   3063 \\
sw       &                 Swahili &  Niger-Congo, Atlantic-Congo, Volta-Congo  &   1810 \\
ts       &                Xitsonga &  Niger-Congo, Atlantic-Congo, Volta-Congo  &    804 \\
twi      &                     Twi &  Niger-Congo, Atlantic-Congo, Volta-Congo  &   3481 \\
yo       &                  Yoruba &  Niger-Congo, Atlantic-Congo, Volta-Congo  &   8522 \\
\bottomrule
\end{tabular}
\caption{The languages in the training dataset, with language families and length of training splits in the dataset. The family classification is our abbreviation of data gathered from the Ethnologue dataset \citep{ETHNOdata} and from Wikipedia. This classification is merely a functional grouping to apply to the task at hand, and not assumed to be authoritative.  }
\label{tab:languages}
\end{table*}

\begin{table*} % report_results.ipynb

\begin{tabular}{lrrrl} % results/save_results-21feb/feb21results.txt
\toprule
\textbf{train-category} & \textbf{in-language} & \textbf{language-cat} & \textbf{multilingual} & \textbf{Comment} \\
\textbf{test-language}          &             &              &              &         \\
\midrule
MoroccanArabic/Darija, &       \textbf{97.5\%} &        96.2\% &        96.7\% &    Arabic     \\
Igbo                   &       \textbf{78.8\%} &        77.6\% &        78.5\% &    Train size     \\
Hausa                  &       \textbf{77.7\%} &         NA &        76.2\% &    Train size     \\
Yoruba                 &       71.3\% &        \textbf{71.4\%} &        70.0\% &    Train size     \\
Mozambican Portuguese   &       \textbf{71.0\%} &       \textbf{ 71.0\%} &        68.8\% &     Portuguese    \\
Algerian Arabic         &       \textbf{68.1\%} &        66.4\% &        61.2\% &       Arabic  \\
Kinyarwanda            &       \textbf{60.9\%} &        57.1\% &        55.6\% &         \\
Amharic                &       \textbf{59.9\%} &        56.2\% &        57.1\% &         \\
Twi                    &       \textbf{58.7\%} &        56.8\% &        56.7\% &         \\
Xitsonga               &       \textbf{54.9\%} &        50.1\% &        45.9\% &         \\
Nigerian Pidgin        &       51.1\% &        \textbf{51.8\%} &        50.1\% &         \\
Swahili                &       \textbf{50.5\%} &        49.4\% &        46.0\% &         \\
\bottomrule
\end{tabular}
\caption{\fen-scores from subsequent experiments after submission. The \textit{XLMT-sentiment} model was fine-tuned on either the one language tested only (In-language), the combined training data from the languages in the target model's language family (language-cat), or on the complete multilingual dataset. We find that the best performing models are either trained on the languages with the largest training dataset, or on languages related to languages that were seen both during model pretraining and initial fine-tuning. Best result for each language is printed in boldface.}
\label{tab:train-data}
\end{table*}

% \begin{figure*}
%     \centering
%     \includegraphics{"img/multi-mono.png"}
%     \caption{\textbf{Improvement from training on multilingual data, as compared with monolingual.} Grey columns are model performance when fine-tuned on monolingual training data. Blue line is relative performance of a model fine-tuned on all languages in the language's family. Orange dotted line is relative performance of models fine-tuned on the multilingual dataset, as compared to the model finetuned on the target language only. The numbers are shown in Table \ref{tab:train-data}. }
%     \label{fig:relative-multi}
% \end{figure*}

\section{Our submission}\label{seq:ourwork}
Our self-imposed constraints on the experiments have been:
\begin{itemize}
    \item Use no language data outside the provided datasets
    \item Use no pretrained llm larger than \textit{"base"} size
    \item No additional pretraining of the llm
    \item Use the same llm for fine-tuning on all languages
\end{itemize}

Our experiments have sought to answer two questions: 
\begin{enumerate}
    \item[a)] What pretrained llm can be a good base for sentiment analysis in the provided low-resource languages?
    \item[b)] Can we combine data for the provided languages to provide a training set that performs better than the single-language dataset?
\end{enumerate}

\paragraph{Our answer to question a)} is found in Section \ref{sec:plm} and Table \ref{tab:init_models} where we conclude that XLMT-sentiment is our best model to fine-tune for these languages. 

\paragraph{To answer b)} we test all languages on the model fine-tuned on the multilingual dataset prepared for subtask B. We also create subsets of languages based on language families or classifications. We decide on the subsets of Afro-Asiatic-Semitic, Volta-Congo, and Creole. the groupings were derived from information in the Ethnologue dataset \citep{ETHNOdata} and from Wikipedia.\footnote{\url{https://www.wikipedia.org/}}. Hausa was the only Chadic language in the training data, and was not evaluated against any language family dataset. Our reasoning for evaluating each language against multilingual training data, is that since some of the languages are poorly related to data used in the pretraining of the llm, more data may be better. But due to the "curse of multilinguality" \citep{conneau-etal-2020-unsupervised} where it is observed that adding more and more languages comes at a cost, we also speculate that training only on languages within the same language family might help.

 During the initial experiments that lead to our choices for the competition submission, we found that only for Swahili did the model perform better when being fine-tuned on the multilingual dataset, than when being fine-tuned on its own language's training data. Our submission for Swahili is therefore based on a model fine-tuned on the multilingual dataset, while for all the other languages, their monolingual datasets were used.
\subsection{Our chosen pretrained language model} \label{sec:chosen_model}
The XLMT-sentiment language model was introduced in Section \ref{sec:plm}. The XLMT-sentiment language model was fine-tuned on a dataset for sentiment classification on eight different languages, including Arabic, English and Portugese. Thus, the model was already fine-tuned for the task at hand. Our fine-tuning is therefore a subsequent fine-tuning for the same task, but with data from other languages. 
Due to resource constraints, we used the base version of all models, no large version. XLMT-sentiment\textsubscript{base} is, apart from the classification head, a further trained version of XLM-Roberta\textsubscript{base}. The XLM-Roberta models were trained with a Sentence Piece (SPM) tokenizer. A few other details on the architecture are presented in Table \ref{tab:xlmr}: 

\begin{table}[h]
    \centering
    \begin{tabular}{l r}
    \toprule
        \textbf{Detail} &  \textbf{Value}\\
    \midrule
         Languages & 100 \\
         Vocabulary & 250K \\
         Layers & 12 \\
         Parameters & 270M \\
    \bottomrule     
    \end{tabular}
    \caption{A few details on the XLM-Roberta\textsubscript{base} llm \citep{conneau-etal-2020-unsupervised}. This model was further trained and fine-tuned into XLMT-sentiment, the model chosen for our contribution.  }
    \label{tab:xlmr}
\end{table}

\subsection{ Hyperparmaters for fine-tuning}
All fine-tuning experiments are performed with a Huggingface AutoModelForSequenceClassification wrapper around the pretrained llm.
For the competition contribution, we concatenated the labelled train- and dev-data for each language and for the multilingual dataset.

The only hyperparameters we searched for, were the amount of epochs to train, within the maximum of seven epochs. The epochs selected for each single-language model were:\\ 
\verb|dz:7, am:5, yo:6, twi:4, pcm:6,pt:7|
\verb| ma:7, ha:4, ig:6, ts:5, kr:7|\\ 
The symbol for each language is found in Table \ref{tab:languages}.
A few other hyperparameters are found in Table \ref{tab:finetune_params}:
\begin{table}[h]
    \centering
    \begin{tabular}{l r}
    \toprule
        \textbf{Hyper-parameter} & \textbf{Value} \\
    \midrule
       Learning rate  & 2e-5 \\
       Warmup-steps & 100 \\
       Weight decay & 0.01 \\
       Batch size & 32 \\
    \bottomrule
    \end{tabular}
    \caption{A few details on our hyper-parameters for fine-tuning our llms on the Afrisenti datsets.}
    \label{tab:finetune_params}
\end{table}

\subsection{Competition results}
Our results in the competition were around average or lower. Taking into account our constraint on llm size and on the fact that no other target language resources were applied, we find the results reasonable. Our code will be available on github.\footnote{\url{https://github.com/egilron/AfriSenti-SemEval-2023}}

\subsection{Subsequent analysis}
After our submission to the competition, we re-ran the experiments, fine-tuning on the training split, and evaluating on the labelled development split. Table \ref{tab:train-data} reports the findings from these experiments, where we allowed the model to train for up to  14 epochs. Under these new conditions we see that Swahili would also benefited from inference on a model fine-tuned on its own training data only.
 
 Table \ref{tab:train-data} shows that nearly all languages had better results fine-tuning only on their own language. We believe that the fact that virtually all languages hava training samples in the thousands, gives the model enough in-language signal, and that the added data from other languages adds too much noise. This is in line with our earlier findings where we for a lower-resourced language, found that adding related English data was mostly beneficial only when the in-language samples were less than 500 \citep{ronningstad2020targeted}.  \\

\section{Conclusion}
We have shown how the \textit{Twitter-xlmr-sentiment} model can be a helpful resource and starting point for sentiment analysis in low-resource languages. We have seen that fine-tuning with a multilingual dataset was in general not helpful for these language data, with training samples in the thousands. A suggestion for further work is to fine-tune models with only ten, or a hundred in-language training samples, and measure the value of adding multilingual data in those few-shot situations.

We have found that best results were achieved for languages that either have the largest training set, or what we assume are languages close to higher resourced languages that have been seen during training and initial fine-tuning. We find that Nigerian Pidgin performed second to worst. We were expecting this language to perform better due to its supposedly relatedness to English. We have not attempted to quantify any language similarities, and have no explanation why Nigerian Pidgin performed so poorly.

\section{Ethical considerations}
In this work we are performing experiments on several low-resource African languages. Our intent is to learn from this language diversity, and contribute towards a stronger digital presence for these languages. This can be viewed as giving people stuff they have not asked for, as we do not know to what degree this is a felt need among the actual language communities. But we also consider all languages to be worth studying and learning from, whether or not this study is of immediate experienced benefit to the language users or not. We are therefore thankful to the organizers for allowing us to work on these languages, and we do not assume that our work is of direct benefit to others than ourselves. We have only conducted work that we ourselves appreciate, when others conduct similar work on our own not-so-highly resourced native language.
\section*{Acknowledgements}
The work documented in this publication has been carried out within the NorwAI Centre for Research-based Innovation, funded by the Research Council of Norway (RCN), with grant number 309834.
% , and the SANT project (Sentiment Analysis for Norwegian Text), also funded by the RCN (grant number 270908). 
The computations were performed on resources provided by UNINETT Sigma2 - the National Infrastructure for High Performance Computing and Data Storage in Norway.

% \citet{muhammad-etal-2022-naijasenti} har noe på hvilke nigerianske språk som er med i pretraining xlmr

\bibliography{anthology,custom}

\begin{thebibliography}{13}
\expandafter\ifx\csname natexlab\endcsname\relax\def\natexlab#1{#1}\fi

\bibitem[{Agarwal et~al.(2011)Agarwal, Xie, Vovsha, Rambow, and
  Passonneau}]{agarwal-etal-2011-sentiment}
Apoorv Agarwal, Boyi Xie, Ilia Vovsha, Owen Rambow, and Rebecca Passonneau.
  2011.
\newblock \href {https://aclanthology.org/W11-0705} {Sentiment analysis of
  {T}witter data}.
\newblock In \emph{Proceedings of the Workshop on Language in Social Media
  ({LSM} 2011)}, pages 30--38, Portland, Oregon. Association for Computational
  Linguistics.

\bibitem[{Alabi et~al.(2022)Alabi, Adelani, Mosbach, and
  Klakow}]{alabi-etal-2022-adapting}
Jesujoba~O. Alabi, David~Ifeoluwa Adelani, Marius Mosbach, and Dietrich Klakow.
  2022.
\newblock \href {https://aclanthology.org/2022.coling-1.382} {Adapting
  pre-trained language models to {A}frican languages via multilingual adaptive
  fine-tuning}.
\newblock In \emph{Proceedings of the 29th International Conference on
  Computational Linguistics}, pages 4336--4349, Gyeongju, Republic of Korea.
  International Committee on Computational Linguistics.

\bibitem[{Aue and Gamon(2005)}]{aue2005customizing}
Anthony Aue and Michael Gamon. 2005.
\newblock \href
  {https://www.microsoft.com/en-us/research/publication/customizing-sentiment-classifiers-to-new-domains-a-case-study/}
  {Customizing sentiment classifiers to new domains: a case study}.
\newblock In \emph{Submitted to RANLP-05, the International Conference on
  Recent Advances in Natural Language Processing}.

\bibitem[{Barbieri et~al.(2022)Barbieri, Espinosa~Anke, and
  Camacho-Collados}]{barbieri-etal-2022-xlm}
Francesco Barbieri, Luis Espinosa~Anke, and Jose Camacho-Collados. 2022.
\newblock \href {https://aclanthology.org/2022.lrec-1.27} {{XLM}-{T}:
  Multilingual language models in {T}witter for sentiment analysis and beyond}.
\newblock In \emph{Proceedings of the Thirteenth Language Resources and
  Evaluation Conference}, pages 258--266, Marseille, France. European Language
  Resources Association.

\bibitem[{Conneau et~al.(2020)Conneau, Khandelwal, Goyal, Chaudhary, Wenzek,
  Guzm{\'a}n, Grave, Ott, Zettlemoyer, and
  Stoyanov}]{conneau-etal-2020-unsupervised}
Alexis Conneau, Kartikay Khandelwal, Naman Goyal, Vishrav Chaudhary, Guillaume
  Wenzek, Francisco Guzm{\'a}n, Edouard Grave, Myle Ott, Luke Zettlemoyer, and
  Veselin Stoyanov. 2020.
\newblock \href {https://doi.org/10.18653/v1/2020.acl-main.747} {Unsupervised
  cross-lingual representation learning at scale}.
\newblock In \emph{Proceedings of the 58th Annual Meeting of the Association
  for Computational Linguistics}, pages 8440--8451, Online. Association for
  Computational Linguistics.

\bibitem[{Ethnologue()}]{ETHNOdata}
Ethnologue. 2017.
\newblock \href {http://www.ethnologue.com} {Ethnologue: Languages of the
  world. global dataset}.

\bibitem[{Liu(2017)}]{liu_2017}
Bing Liu. 2017.
\newblock \emph{Sentiment analysis: mining opinions, sentiments, and emotions}.
\newblock Cambridge University Press.

\bibitem[{Muhammad et~al.(2023{\natexlab{a}})Muhammad, Abdulmumin, Ayele,
  Ousidhoum, Adelani, Yimam, Ahmad, Beloucif, Mohammad, Ruder, Hourrane,
  Brazdil, Ali, David, Osei, Bello, Ibrahim, Gwadabe, Rutunda, Belay, Messelle,
  Balcha, Chala, Gebremichael, Opoku, and Arthur}]{muhammad2023afrisenti}
Shamsuddeen~Hassan Muhammad, Idris Abdulmumin, Abinew~Ali Ayele, Nedjma
  Ousidhoum, David~Ifeoluwa Adelani, Seid~Muhie Yimam, Ibrahim~Sa'id Ahmad,
  Meriem Beloucif, Saif~M. Mohammad, Sebastian Ruder, Oumaima Hourrane, Pavel
  Brazdil, Felermino Dário Mário~António Ali, Davis David, Salomey Osei,
  Bello~Shehu Bello, Falalu Ibrahim, Tajuddeen Gwadabe, Samuel Rutunda, Tadesse
  Belay, Wendimu~Baye Messelle, Hailu~Beshada Balcha, Sisay~Adugna Chala,
  Hagos~Tesfahun Gebremichael, Bernard Opoku, and Steven Arthur.
  2023{\natexlab{a}}.
\newblock \href {https://doi.org/10.48550/arXiv.2302.08956} {{AfriSenti: A
  Twitter Sentiment Analysis Benchmark for African Languages}}.

\bibitem[{Muhammad et~al.(2023{\natexlab{b}})Muhammad, Abdulmumin, Yimam,
  Adelani, Ahmad, Ousidhoum, Ayele, Mohammad, Beloucif, and
  Ruder}]{muhammadSemEval2023}
Shamsuddeen~Hassan Muhammad, Idris Abdulmumin, Seid~Muhie Yimam, David~Ifeoluwa
  Adelani, Ibrahim~Sa'id Ahmad, Nedjma Ousidhoum, Abinew~Ali Ayele, Saif~M.
  Mohammad, Meriem Beloucif, and Sebastian Ruder. 2023{\natexlab{b}}.
\newblock {SemEval-2023 Task 12: Sentiment Analysis for African Languages
  (AfriSenti-SemEval)}.
\newblock In \emph{Proceedings of the 17th {{International Workshop}} on
  {{Semantic Evaluation}} ({{SemEval-2023}})}. {Association for Computational
  Linguistics}.

\bibitem[{Muhammad et~al.(2022)Muhammad, Adelani, Ruder, Ahmad, Abdulmumin,
  Bello, Choudhury, Emezue, Abdullahi, Aremu, Jorge, and
  Brazdil}]{muhammad-etal-2022-naijasenti}
Shamsuddeen~Hassan Muhammad, David~Ifeoluwa Adelani, Sebastian Ruder,
  Ibrahim~Sa{'}id Ahmad, Idris Abdulmumin, Bello~Shehu Bello, Monojit
  Choudhury, Chris~Chinenye Emezue, Saheed~Salahudeen Abdullahi, Anuoluwapo
  Aremu, Al{\'\i}pio Jorge, and Pavel Brazdil. 2022.
\newblock \href {https://aclanthology.org/2022.lrec-1.63} {{N}aija{S}enti: A
  nigerian {T}witter sentiment corpus for multilingual sentiment analysis}.
\newblock In \emph{Proceedings of the Thirteenth Language Resources and
  Evaluation Conference}, pages 590--602, Marseille, France. European Language
  Resources Association.

\bibitem[{Reimers and Gurevych(2019)}]{reimers-gurevych-2019-sentence}
Nils Reimers and Iryna Gurevych. 2019.
\newblock \href {https://doi.org/10.18653/v1/D19-1410} {Sentence-{BERT}:
  Sentence embeddings using {S}iamese {BERT}-networks}.
\newblock In \emph{Proceedings of the 2019 Conference on Empirical Methods in
  Natural Language Processing and the 9th International Joint Conference on
  Natural Language Processing (EMNLP-IJCNLP)}, pages 3982--3992, Hong Kong,
  China. Association for Computational Linguistics.

\bibitem[{R{\o}nningstad(2020)}]{ronningstad2020targeted}
Egil R{\o}nningstad. 2020.
\newblock Targeted sentiment analysis for norwegian text.
\newblock Master's thesis, University of Oslo.

\bibitem[{Yimam et~al.(2020)Yimam, Alemayehu, Ayele, and
  Biemann}]{yimam-etal-2020-exploring}
Seid~Muhie Yimam, Hizkiel~Mitiku Alemayehu, Abinew Ayele, and Chris Biemann.
  2020.
\newblock \href {https://doi.org/10.18653/v1/2020.coling-main.91} {Exploring
  {A}mharic sentiment analysis from social media texts: Building annotation
  tools and classification models}.
\newblock In \emph{Proceedings of the 28th International Conference on
  Computational Linguistics}, pages 1048--1060, Barcelona, Spain (Online).
  International Committee on Computational Linguistics.

\end{thebibliography}
\bibliographystyle{acl_natbib}

% \appendix

% \section{Example Appendix}
% \label{sec:appendix}

% This is a section in the appendix.

\end{document}